\documentclass[letterpaper, 10 pt, conference]{ieeeconf}  

\IEEEoverridecommandlockouts                              

\IEEEoverridecommandlockouts

\usepackage{cite}
\usepackage{amsmath,amssymb,amsfonts}
\usepackage{algorithmic}
\usepackage{graphicx}
\usepackage{textcomp}
\usepackage{subcaption}
\usepackage{xcolor}
\usepackage{hyperref}
\hypersetup{
    colorlinks=true,
    linkcolor=black,
    urlcolor=blue,
    citecolor=black,
    pdfborder={0 0 0}
}

\title{\LARGE \bf
Bioinspired SLAM Approach for Unmanned Surface Vehicle 
}

\author{Fabio Coelho$^{1}$, Joao Victor T. Borges$^{1}$, Paulo Padrao$^{2}$, Jose Fuentes$^{3}$, Ramon R. Costa$^{1}$, Liu Hsu$^{1}$ \\ and Leonardo Bobadilla$^{3}$%
\thanks{Fabio Coelho, Joao Victor T. Borges, Ramon Romankevicius, and Liu Hsu are with the Department of Electrical Engineering, Federal University of Rio de Janeiro, Brazil
        {\tt\footnotesize f.coelho@gsuite.iff.edu.br, \{borgesjvt, ramonrcosta\}@gmail.com, lhsu@coppe.ufrj.br}}%
\thanks{$^{2}$ Paulo Padrao is with the Department of Mathematics and Computer Science, Providence College, Providence, RI, USA
{\tt\footnotesize ppadraol@providence.edu}}%
\thanks{$^{3}$ Jose Fuentes and Leonardo Bobadilla are with the School of Computing and Information Sciences, Florida International University, Miami, FL, USA
{\tt\footnotesize jfuen099@fiu.edu, bobadilla@cs.fiu.edu}}
}

\begin{document}

\maketitle
\thispagestyle{empty}
\pagestyle{empty}

\begin{abstract}
This paper presents OpenRatSLAM2, a new version of OpenRatSLAM—a bioinspired SLAM framework based on computational models of the rodent hippocampus. OpenRatSLAM2 delivers low-computation-cost visual-inertial based SLAM, suitable for GPS-denied environments. Our contributions include a ROS2-based architecture, experimental results on new waterway datasets, and insights into system parameter tuning. This work represents the first known application of RatSLAM on USVs. The estimated trajectory was compared with ground truth data using the Hausdorff distance. The results show that the algorithm can generate a semimetric map with an error margin acceptable for most robotic applications.
\end{abstract}

\section{Introduction}

The increasing use of unmanned surface vehicles (USVs) for scientific, military, and commercial purposes requires the development of robust navigation systems \cite{b1}. Common applications include oceanographic data collection, oil and gas exploration, environmental surveys, mine countermeasures, and surveillance \cite{newaz2023, puthumanaillam2025}. To autonomously perform such tasks, a mobile robot must be able to localize itself within its environment \cite{b2}.

Common approaches include combining GPS with an inertial measurement unit (IMU) and Kalman filtering algorithms for state estimation in USVs \cite{b3}–\cite{b4}. However, these methods fail in GPS-denied environments where satellite signals are obstructed \cite{b5}. Moreover, GPS signals are vulnerable to various disruptions and cyberattacks, including jamming and spoofing \cite{b6}.

To address these limitations, Simultaneous Localization and Mapping (SLAM) is an alternative that enables a vehicle to build a map of its surroundings while estimating its position relative to that map. Many existing SLAM implementations rely on computationally intensive sensors, such as LiDAR or depth cameras. These sensors often require high processing and storage demands, making them less suitable for real-time applications on resource-constrained platforms \cite{b7}.

Motivated by recent advances in neuroscience, several brain-inspired SLAM systems have been proposed \cite{b8}. A pioneering work is the RatSLAM framework, a biologically inspired SLAM algorithm based on computational models of the rodent hippocampus. RatSLAM employs a Continuous Attractor Neural Network (CANN) to construct a cognitive map of an environment using only a low-resolution monocular camera \cite{b9}. Compared to probabilistic SLAM approaches, RatSLAM offers reduced computational complexity and efficient memory usage and is well-suited for both indoor and large-scale outdoor mapping.

In recent years, several RatSLAM-based variants have been proposed \cite{b10}. For instance, \cite{b11} introduced a MATLAB-based RatSLAM implementation in a rat robot, demonstrating its capability to learn spatial layouts. However, the system's performance was too slow for real-time operation in large environments. Another approach, OpenRatSLAM, was proposed as an open-source RatSLAM implementation based on the Robot Operating System (ROS) \cite{b12}. This version benefits from ROS's node parallelization and modular integration with diverse robotic architectures \cite{b10}.

The emergence of ROS 2 as the dominant middleware for new robotic systems has created integration challenges, as OpenRatSLAM was primarily developed for ROS 1. In this context, xRatSLAM was developed as an extensible, parallel, open-source framework implemented as a C++ library to facilitate the development and testing of RatSLAM algorithm variants \cite{b10}.

While most applications have targeted ground robots, RatSLAM-inspired algorithms have also been explored in other domains. One aerial application, NeuroSLAM, is a neuro-inspired SLAM system with four degrees of freedom (4DoF), based on computational models of 3D grid cells and multilayered head direction cells. It integrates visual and self-motion cues through a dedicated vision system \cite{b14}. In underwater environments, two RatSLAM-based systems have been developed: DolphinSLAM \cite{b15}, a 3D variant, and a more recent system that implements Pose Cells using Spiking Neural Networks (SNNs) \cite{b16}. Both were developed using ROS 1 distributions, which are now deprecated and unsupported.

In summary, the contributions of this work are as follows:
\begin{itemize}
\item A new version of OpenRatSLAM, implemented using ROS 2 Rolling, referred to as OpenRatSLAM2. This version benefits from ROS 2's advantages, including improved maintainability and easier integration with modern tools. Additionally, the communication middleware is more robust than ROS 1, providing streamlined transition from simulation to physical robot deployment;
\item To the best of our knowledge, this is the first application of RatSLAM to a USV;
\item A visual-inertial dataset collected using a USV for evaluating SLAM performance in aquatic environments.
\end{itemize}

\section{Problem Formulation}

\begin{figure}[t]
\centerline{\includegraphics[width=1\linewidth]{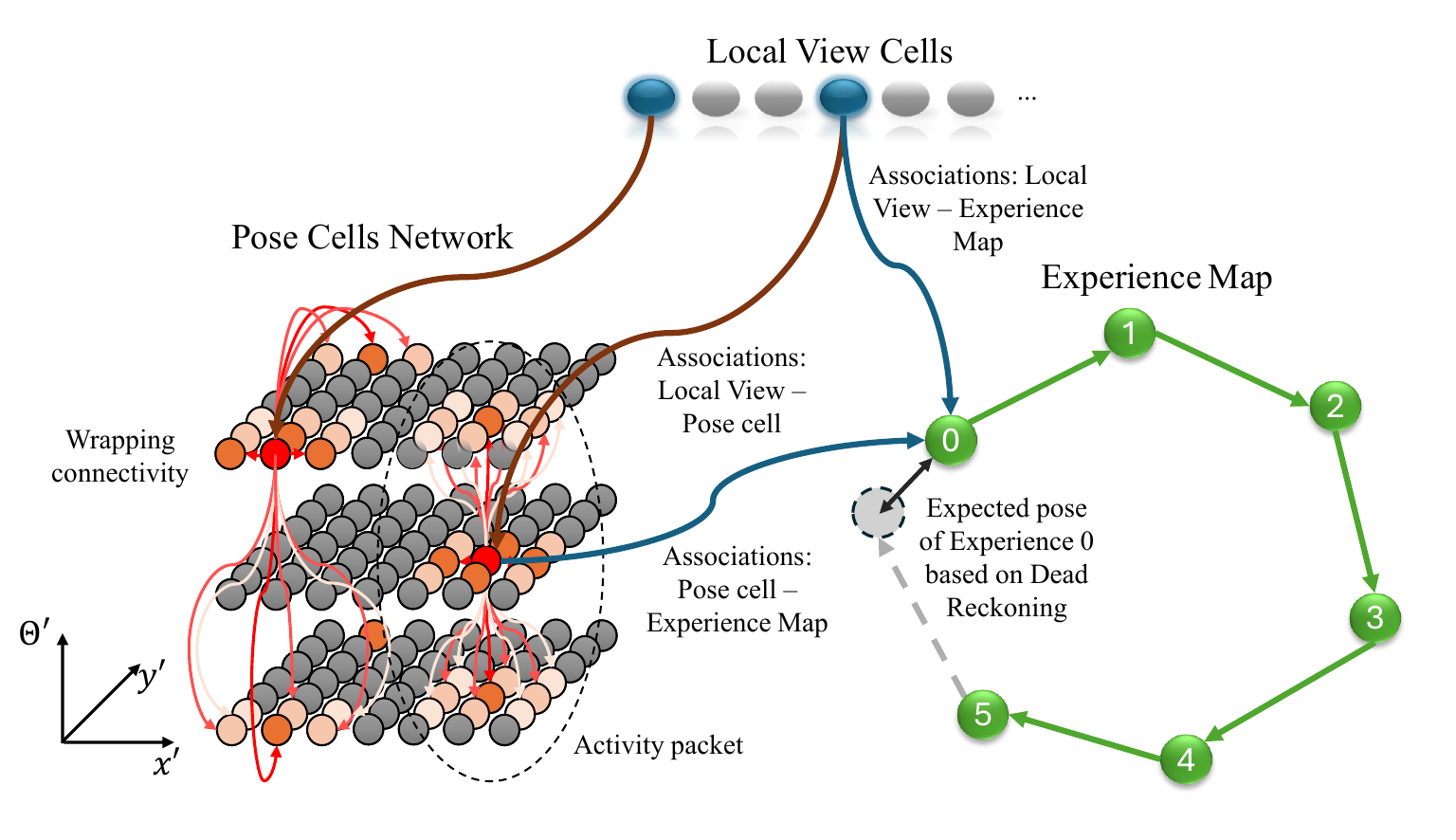}}
\caption{RatSLAM main modules.}
\label{fig}
\end{figure}

RatSLAM is an appearance-based system introduced in \cite{b17} that relies on visual similarity between images captured at discrete locations in the environment. It is a mapping and localization algorithm inspired by the neural processes associated with spatial navigation in the hippocampus and entorhinal cortex of rodents \cite{b18}. Figure~\ref{fig} illustrates the OpenRatSLAM architecture, which builds upon the original RatSLAM framework. The system consists of three main modules: \textit{Local View Cells}, \textit{Pose Cells}, and the \textit{Experience Map}.

The \textit{Local View Cells} are visual templates that represent unique scenes learned in the environment. The \textit{Pose Cells Network} is the core of the algorithm and it models the behavior of three types of cells found in the rodent brain, strongly linked to spatial location: \textit{place}, \textit{head} and \textit{grid} cells.

\textit{Place cells} fire at their peak when the rodent is in a specific location, with excitation decreasing as the animal moves away from that location \cite{b9}. \textit{Head direction cells} activate only when the animal is oriented toward specific global directions \cite{b18_2}, \cite{b18_3}.

The activity packet in the pose cell network encodes the belief about the current pose, denoted by the vector $ \left [ \begin{matrix} x & y &  \theta \\ \end{matrix} \right ]^T$. Each local view cell is anchored, at the time of its creation, to the centroid of an active place cell packet. This association is indicated by the brown lines in Fig.~\ref{fig}. 

The \textit{Experience Map} is a topological graph-based representation that integrates information from pose cells and local view cells. Each node, or experience, stores the estimated robot pose, derived from the centroid of the activation packet, and the corresponding local view ID, both captured at the time of experience creation. These associations are illustrated by the blue lines in Fig.~\ref{fig}.

Over time, odometry-based dead reckoning accumulates drift. For instance, a gray node in the experience map may represent the estimated pose based on odometry alone, while a matching visual template indicates the scene corresponds to a previously observed location (e.g., experience zero). This triggers the loop closure process, which is described in subsection \ref{Subsection II-C}.

\subsection{Pose Cells Network}

Pose Cells Network (PCN) is a continuous attractor network (CAN) implemented as a three-dimensional structure of cells with weighted excitatory and inhibitory connections \cite{b19}. It exhibits characteristics similar to those of navigation-related neurons found in several mammalian brains, particularly grid cells. Cells within the network are locally linked via excitatory connections that wrap around all six faces of the network. Additionally, each cell inhibits all other cells in the network, which contribute to the formation of localized activity packets. \cite{b12}. Unlike most artificial neural networks, the CAN model does not update its state by adjusting connection weights. Instead, it is updated by varying the activity levels of its neural units\cite{b19}.

\subsubsection{Attractor Dynamics}

The intrinsic attractor dynamics of the network ensure that, in the absence of external input, the activity converges over several iterations to a single localized packet. For each pose cell, \textit{local excitation} is achieved through a three-dimensional Gaussian distribution of weighted connections. This process is visually represented by the red arrows in Fig. ~\ref{fig}. In the OpenRatSLAM implementation, the excitatory weight matrix $\varepsilon$ is given by Eq. \eqref{excitation_gauss}
\begin{equation}
\varepsilon_{a, b, c} = \dfrac{1}{\sigma_\varepsilon \sqrt{2 \pi}} \text{e}^ {\left ( \frac{-(a-x_c)^2(b-y_c)^2(c-\theta_c)^2}{2 \sigma_\varepsilon ^2}\right )},
\label{excitation_gauss}
\end{equation}
where $a$, $b$, and $c$ represent the distances between cells in $x'$, $y'$ and $\theta'$ coordinates, respectively; $\sigma_\varepsilon$ denotes the variance of the excitation kernel; and $\left [ \begin{matrix} x_c & y_c &  \theta_c \\ \end{matrix} \right ]^T$ is the center of the activity distribution.

The resulting change in the activity of a pose cell due to local
excitation is given by Eq. \eqref{excitation_step}
\begin{equation}
 \Delta P_{x',y',\theta'} = \sum_{i=0}^{n_{x'}-1}\sum_{j=0}^{n_{y'}-1}\sum_{k=0}^{n_{\theta'}-1} P_{i,j,k} \varepsilon_{a,b,c},
\label{excitation_step}
\end{equation}
where $n_{x'}$, $n_{y'}$ e $n_{\theta'}$ denote the dimensions of the pose cell matrix in units of cells. Equation \ref{excitation_step} represents a circular convolution of two three-dimensional matrices.

The following steps are \textit{local} and \textit{global inhibition}. Each cell inhibits nearby cells using an inhibitory kernel $\psi_{a,b,c}$, which is a broader version the excitation kernel. The variances for inhibition are larger than for excitation, creating the so-called Mexican-hat function. Furthermore, a constant global inhibition $\varphi$ is uniformly applied across all cells. The total inhibitory update is described by Eq. \eqref{inhibition_step}
\begin{equation}
    \Delta P_{x',y',\theta'} = -\sum_{i=0}^{n_{x'}-1}\sum_{j=0}^{n_{y'}-1}\sum_{k=0}^{n_{\theta'}-1} P_{i,j,k} \psi_{a,b,c} - \varphi,
    \label{inhibition_step}
\end{equation}
where $\varphi$ is the global inhibition constant. After inhibition, all pose cells values $P$ are clipped to be nonnegative and then \textit{normalized} so that the total energy in the network sums to one \cite{b9}.

\subsubsection{Path Integration}

In the RatSLAM system, path integration consists of shifting the activity packet across the pose cell network based on odometry information. Although this approach presents lower biological fidelity compared to computing transitions through weighted connections, it is computationally efficient and avoids scalability issues. Unlike probabilistic SLAM approaches, this mechanism does not increase uncertainty over time. Odometry data used for path integration can be extracted from image-based motion estimation or obtained from other odometric sources. The accumulated error in the path integration process is reduced by the activation of local view cells, which inject energy into the pose cell network when familiar scenes are detected, thereby enabling loop closure and correction of the robot's estimated pose.

\subsection{Local View Cells}
Local view cells consist of an expandable array of units, denoted by $V$, where each cell encodes a distinct visual scene. A given cell becomes active when the robot perceives its corresponding scene. While there are no direct connections between local view cells themselves, connections are formed between local view cells and pose cells upon creation of a new visual template. When a new local view cell $V_i$ is created, an excitatory link $\beta_i$ is learned between this unit and the centroid of the currently dominant activity packet in the pose cell network. If the same visual scene is encountered again, the associated local view cell injects activity into the pose cells through this excitatory link. This process is described by Eq. \eqref{activity_injection}
\begin{equation}
    \Delta P_{x',y',\theta'} = \delta \sum_{i} \beta_{i, x', y', \theta'} V_i.
    \label{activity_injection}
\end{equation}

where $\delta$ is a constant that determines the influence of visual cues on the correction of robot's pose estimate. A saturation mechanism is implemented to limit the duration for which a visual template can inject activity, preventing erroneous relocations in the absence of movement. Successful relocalization requires the robot to perceive a familiar sequence of images, resulting in a series of energy injections into the same region of the pose cell network. The visual energy injection process is critical, and its performance is sensitive to parameter tuning as provided in Section \ref{sec_parameters}.

\subsection{Experience Map}\label{Subsection II-C}
Although pose cells represent a finite area, the wrapping of the network edges allows an infinite area to be mapped. As a result, a single pose cell may correspond to multiple physical locations. To solve potential ambiguities, the experience map is a semi-metric topological map \cite{b19}, that estimates a unique pose of the robot by combining information from pose cells, local view cells and odometry. The experience map consists of a graph, where each node (referred to as \textit{experience}) is defined as a 3-tuple $e_i =\left\{ P^i, V^i, \mathbf{p}^i\right\}$, where $P^i$ and $V^i$ are the pose cell and local view activity states, respectively, at the time of experience creation. The term $\mathbf{p}^i$ represents the estimated pose of the robot within the coordinate space of the experience map.

\subsubsection{Experience Creation}
The robot's current pose and local view information are compared against all previously stored experiences through a matching score metric $S^i$ provided by Eq. \eqref{matching_score}
\begin{equation}
    S^i = \mu_p \left | P^i - P \right| + \mu_\upsilon \left | V^i - V \right|. 
    \label{matching_score}
\end{equation}
$\mu_p$ and $\mu_\upsilon$ are weighting factors for the pose and local view components, respectively. If the minimum score across all stored experiences satisfies $\min(\mathbf{S}) \geq S_{\text{max}}$, indicating that the current state is sufficiently distinct, a new experience is created.
When a new experience $e_j$ is added to the graph, a transition $l_{ij}$ is also established between $e_j$ and the previously active experience $e_i$. This transition is represented in Eq. \eqref{transition}
\begin{equation}
    l_{ij} =  \left\{ \Delta\mathbf{p}^{ij}, \Delta t^{ij} \right\},
    \label{transition}
\end{equation}
where $\Delta\mathbf{p}^{ij}$ denotes the relative pose change computed from odometry, and $\Delta t^{ij}$ is the elapsed time since the last experience. The new experience $e_j$ is given by Eq. \eqref{new_exp}
\begin{equation}
    e_j =\left\{ P^j, V^j, \mathbf{p}^i + \Delta p^{ij}\right\}.
    \label{new_exp}
\end{equation}

Equation \eqref{new_exp} T is valid only at the time of experience creation. The value of $\mathbf{p}^j$ may subsequently be adjusted during loop closure.

\subsubsection{Loop Closure} 
If any of the stored experience matching scores fall below the threshold $S_\text{max}$, the experience with the lowest score is chosen as the active experience. This experience represents the best estimate of the robot's current location, thereby triggering the loop closure process. At this point, the relative pose between the two matched experiences typically differs from the pose change predicted by odometry. In this case, an \textit{experience map relaxation} procedure is applied. This process minimizes the error between the observed transitions and the absolute poses of experiences within the map. The pose of all experiences are updated using Eq. \eqref{loop_closure}
\begin{equation}
\small
    \Delta \mathbf{p}^i = \alpha \left [ \sum_{j=1}^{N_f} \left ( \mathbf{p}^j 
 - \mathbf{p}^i - \Delta\mathbf{p}^{ij}\right ) + \sum_{k=1}^{N_t} \left ( \mathbf{p}^k 
 - \mathbf{p}^i - \Delta\mathbf{p}^{ki}\right )\right]
    \label{loop_closure},
\end{equation}
where $\alpha$ is a correction rate constant, $N_f$ is the number of links from experience $e_i$ to others, and $N_t$ is the number of links from other experiences to $e_i$. A value of $\alpha = 0.5$ ensures a balance between the velocity of the correction and the stability of the map, while higher values of $\alpha$ may cause instability\cite{b20}. 

\begin{figure*}[]
\centerline{\includegraphics[width=.95\linewidth]{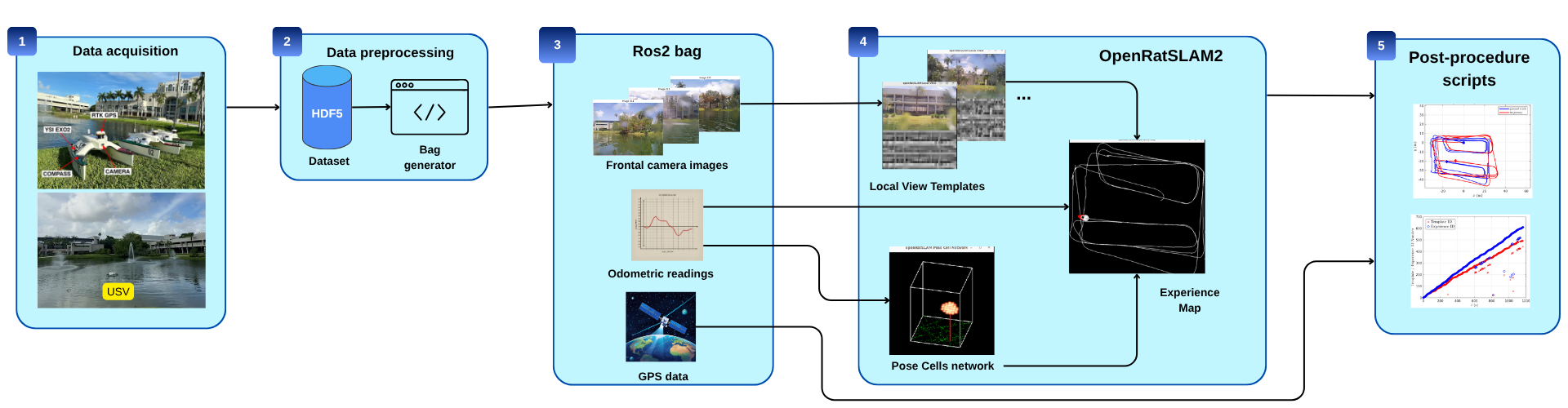}}
\caption{OpenRatSLAM2 workflow: (\textit{i}) Data acquisition while the USV travels a predetermined trajectory; (\textit{ii}) Data preprocessing consists of adapting the input format to the OpenRatSLAM2 framework; (\textit{iii}) A ROS2 bag file containing the data from the sensors of interest; (\textit{iv}) RatSLAM main modules, where input data are processed to generate the experience map; (\textit{v}) Scripts for viewing and analyzing results.}
\label{GraphicalAbtract}
\end{figure*}

\section{System Overview}
In this work, we employed the same data formats used in the first version of OpenRatSLAM. Input data, such as odometric readings and camera images, are read from a ROS bag. Figure \ref{GraphicalAbtract} shows the general workflow. Data collection was performed by the SeaRobotics Surveyor (Fig. ~\ref{GraphicalAbtract}-(i)), and initially stored in Hierarchical Data Format version (HDF5). Next, the HDF5 data file was converted to the ROS 2 bag format to be processed by the OpenRatSLAM2 framework. During OpenRatSLAM2 running, the topic data are recorded into a separate ROS bag file. Output data, such as the generated map, are extracted and visualized using a set of post-processing scripts.$^{\footnotemark}$ 

\footnotetext{\url{https://github.com/OpenRatSLAM2/ratslam}}

Figure ~\ref{GraphicalAbtract} shows the three sensors used to compose the datasets: compass, camera, and GPS. The ISA500 compass includes an integrated Attitude and Heading Reference System (AHRS), with three MEMS-based gyroscopes, accelerometers, and magnetometers and provides the odometric readings. The onboard camera captures frontal-view images,  while the GPS is used to generate the \textit{ground truth}.

\begin{figure}[t]
\centerline{\includegraphics[width=.85\linewidth]{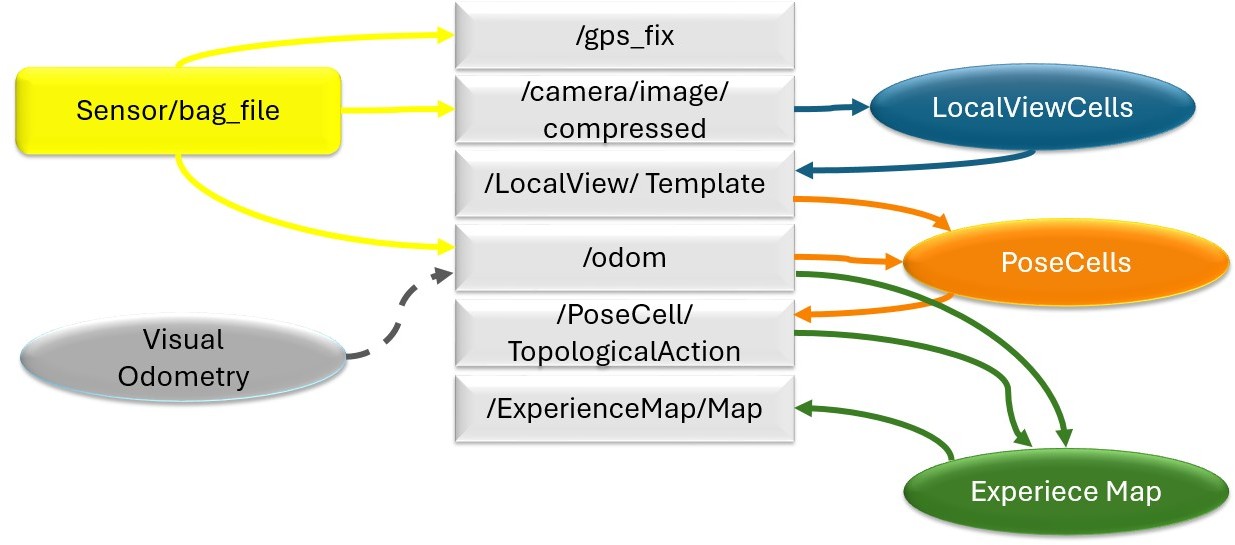}}
\caption{ROS Computational Graph.}
\label{ROSgraph}
\end{figure}

 OpenRatSLAM2 follows the same architectural design as OpenRatSLAM \cite{b12}, which consists of four ROS 2 nodes, as depicted in Fig.~\ref{ROSgraph}:

 \begin{itemize}
     \item \textit{Visual Odometry}: provides an odometry estimate based on image changes. This node not used in this work, as the dataset includes an alternative odometric source;
     \item \textit{Local View Cells}: verifies whether the current view corresponds to a previously encountered scene;
     \item \textit{Pose Cell Network}: the core node of the system; manages the activity packet to estimate pose based on odometric and local view connections. It also handles the experience map nodes and links creation.
     \item \textit{Experience Map}: builds experience graph, performs graph relaxation, and handles path planning.
 \end{itemize}

\section{OpenRatSLAM2 Parameters}\label{sec_parameters}

OpenRatSLAM's performance depends on a large set of parameters. To support reproducibility and facilitate system tuning, this section presents the influence of the main parameters and the expected impact of modifying them. All parameters and their default values can be seen on the project page $^{\footnotemark}$
\footnotetext{\url{{https://openratslam2.github.io/ratslam.github.io/}}}. 

Regarding the Local View module's parameters, \textit{image\_crop} parameters define regions of the image relevant for localization. In this dataset, for example, the water surface can be excluded.  The \textit{template\_size} parameters define the spatial resolution of the template. Small values may fail to capture relevant features, while large values increase memory and computation and may introduce over-sensitivity.

Normalization parameters mitigate the effects of lighting variation. For example, \textit{vt\_normalization} $\approx 1$ maintains the original contrast, adjusting only the brightness. High values $( > 2)$, on the other hand, can cause saturation.

Among the comparison parameters, \textit{vt\_match\_threshold }is critical: lower values (0.01–0.03) reduce false positives and increase the number of templates created—suitable for repetitive environments. Higher values (0.05-0.1) tolerate greater variation, but risk confusing different locations.

Most pose cell parameters do not require frequent tuning. \textit{pc\_dim\_x} sets the spatial resolution of the pose cell network. A larger network improves distinctiveness and loop closure accuracy at the cost of increased computation. \textit{pc\_cell\_x\_size} should match the robot’s speed profile—ideally, the energy packet moves one cell per iteration.

For static environments or slow-moving robots, \textit{vt\_active\_decay} can be increased to 1.5. In dynamic environments or at higher speeds, a value between 0.3 and 0.8 is recommended. The parameter \textit{pc\_vt\_inject\_energy} regulates how strongly visual input corrects pose estimates—0.1–0.2 is typical for stable conditions, while 0.05–0.1 may be better suited for dynamic scenarios. The \textit{exp\_delta\_pc\_threshold} affects the density of the topological map. Lower values $(<1.5)$ generate denser, more detailed maps at the cost of memory. Higher values $(>3.0)$ reduce node creation and computational load, but may reduce environmental resolution. The parameter \textit{exp\_loops} defines how many iterations of the relaxation algorithm are executed per system update.

\section{Experimental Results}

This section presents the results obtained from the experiment conducted using the Green Library Lake dataset, collected at Florida International University (FIU). The dataset includes LiDAR data, odometry, frontal camera images, GPS, and water quality data, all sampled at 1 Hz. The images and odometry data were re-encoded as \texttt{CompressedImage} and \texttt{Odometry} messages, respectively, and stored in a ROS 2 bag file. During preprocessing,  images were resized to 640x480 pixels. To ensure accurate synchronization across data streams, the original timestamps were preserved. Table~\ref{param_table} summarizes the parameter values used in the experiment.

\begin{table}[]
\centering
\footnotesize
\caption{Parameter Values.}
\label{param_table}
\begin{tabular}{lll}
\cline{1-1} \cline{3-3}
\textit{\textbf{Local View}}    &  & \textit{\textbf{Pose Cells}}     \\ \cline{1-1} \cline{3-3} 
image\_crop\_x\_min = 40        &  & pc\_dim\_xy = 18                 \\
image\_crop\_x\_max = 600       &  & pc\_cell\_x\_size = 1            \\
image\_crop\_y\_max = 150       &  & pc\_vt\_inject\_energy = 0.2     \\
image\_crop\_y\_min = 300       &  & exp\_delta\_pc\_threshold = 2.0  \\
\textit{template\_x\_size = 60} &  & vt\_active\_decay = 1.0          \\
template\_y\_size = 20          &  & pc\_vt\_restore = 0.05           \\
vt\_shift\_match = 25           &  &                                  \\ \cline{3-3} 
vt\_step\_match = 5             &  & \textit{\textbf{Experience Map}} \\ \cline{3-3} 
vt\_match\_threshold = 0.073    &  & exp\_loops = 50                  \\
vt\_normalisation = 0           &  & exp\_initial\_em\_deg = 180      \\
vt\_patch\_normalise = 2        &  & exp\_correction = 0.5            \\
vt\_panoramic = 0               &  &                                 
\end{tabular}
\end{table}

\begin{figure}[b]
\centerline{\includegraphics[width=.9\linewidth]{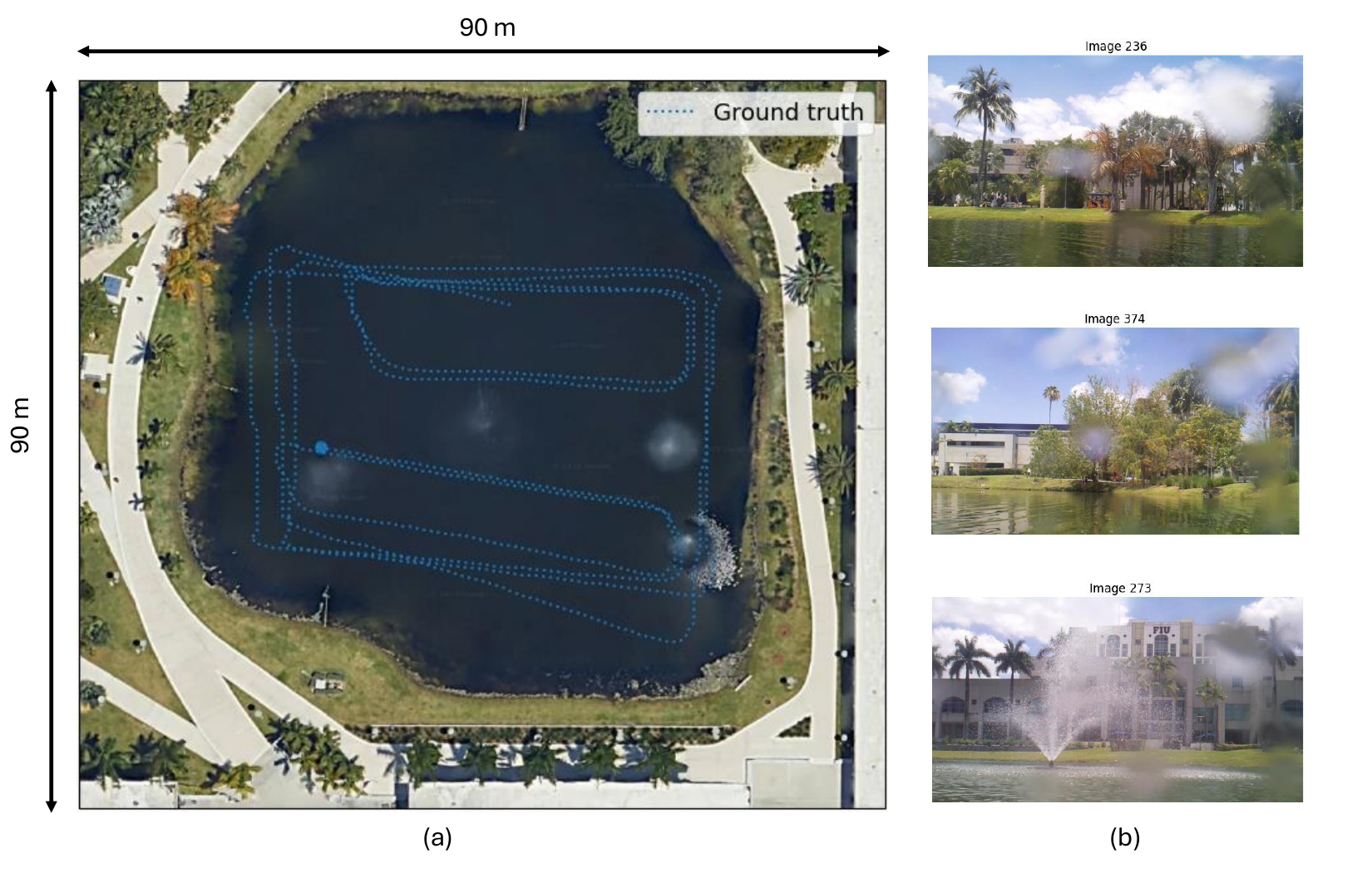}}
\caption{(a) FIU MMC Lake Dataset and (b) frames examples from frontal camera.}
\label{Lake_Dataset}
\end{figure}

\begin{figure}[h!]
\begin{minipage}{0.3\linewidth}
    \centering
    \includegraphics[width=1\linewidth]{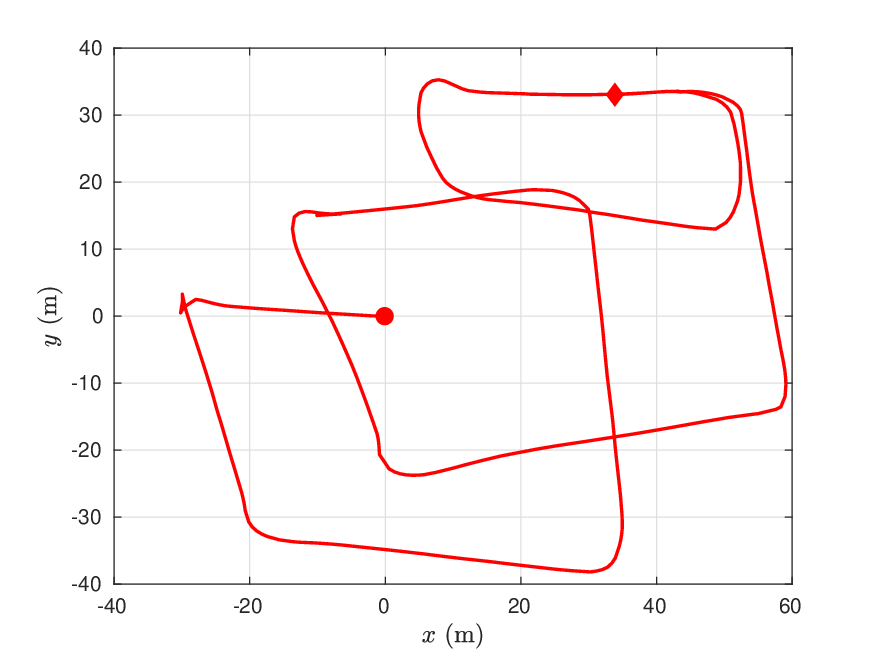}

    \includegraphics[width=1\linewidth]{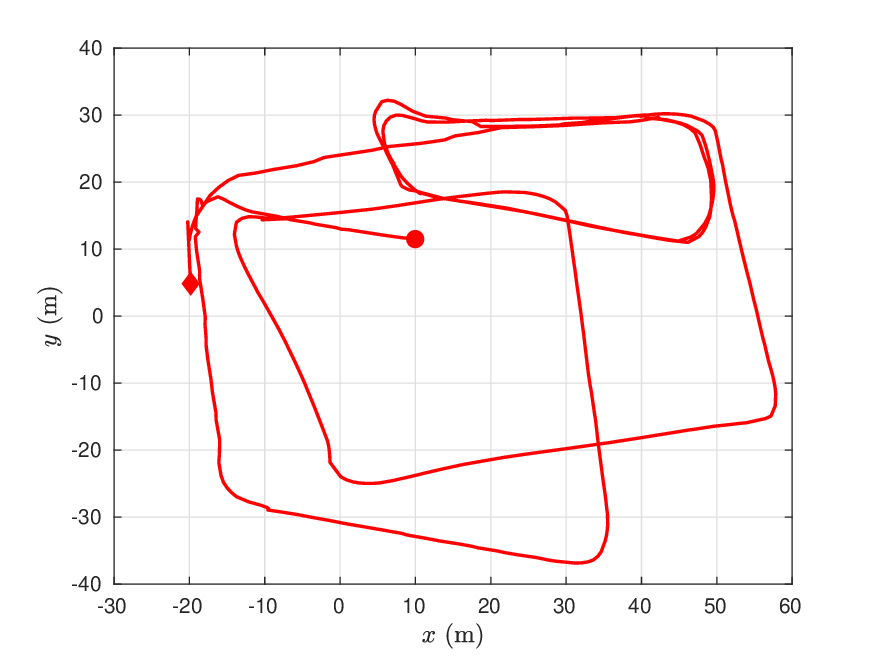}

    \includegraphics[width=1\linewidth]{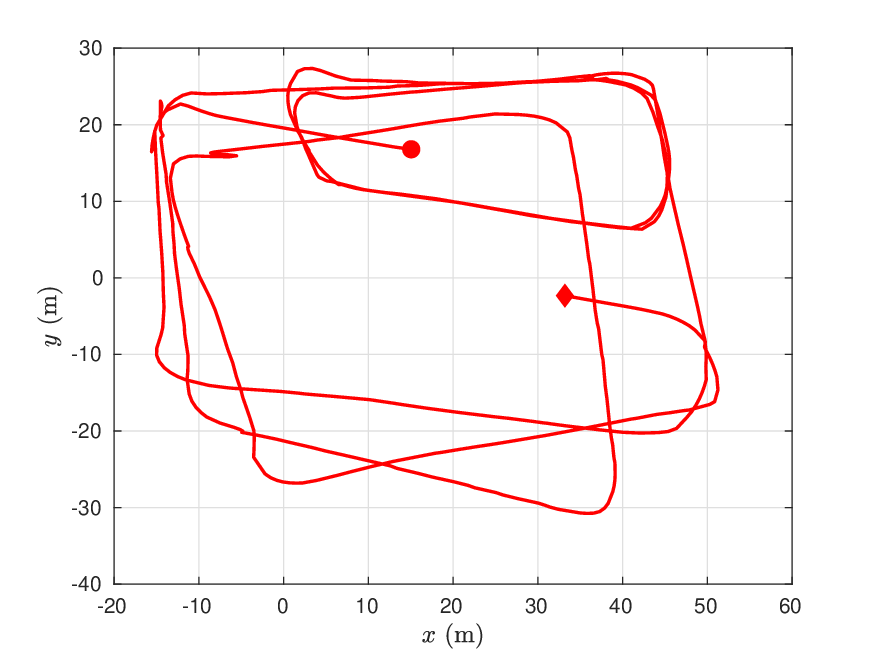}
    \subcaption{}
    \label{fig:map_evo_4}
\end{minipage}
\begin{minipage}{0.65\linewidth}
    \centering
    \includegraphics[width=1\linewidth]{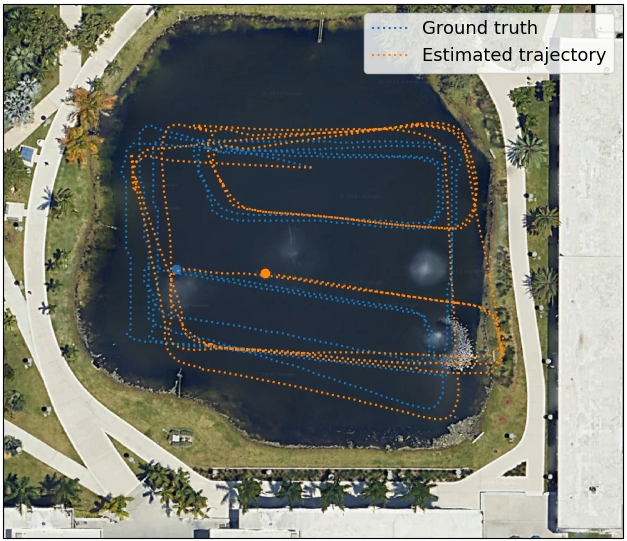}
    \subcaption{}
    \label{fig:map_evo_6}
\end{minipage}
\caption{(a) Experience map evolution over time and (b) final map.}
    \label{fig:em_evolution}
\end{figure}

Figure~\ref{Lake_Dataset} (a) shows the trajectory followed by the USV, which consists of approximately 900 meters over a duration of about 16 min. Figure ~\ref{Lake_Dataset} (b) presents sample frames captured by the onboard camera. (Figs. \ref{fig:map_evo_4}) shows the evolution of the experience map over time. At first, the USV had already completed two full outer loops and one inner loop in the upper region of the lake. At this stage, no loop closure had yet occurred, and odometry drift is visible. In the second figure, a loop closure with the starting region is detected, triggering map convergence toward the ground truth. With subsequent loop closures, the map becomes increasingly stable, requiring only minor refinements.

The final map is shown in Figure \ref{fig:map_evo_6} where the estimated trajectory (orange) closely aligns with the ground truth (blue). A complete visualization of the trajectory evolution is available in the accompanying video $^{\footnotemark}$
\footnotetext{\url{{https://openratslam2.github.io/ratslam.github.io/}}}. The accuracy of the estimated trajectory was evaluated using the \textit{Hausdorff distance}. Given a metric space $(X,d)$ and two non-empty subsets $A, B \subseteq X$, the Hausdorff distance $d_H(A, B)$ is defined as Eq. \eqref{Hausdorff}:

\begin{equation}
    d_H(A, B) = \max\left( 
    \sup_{a \in A} \inf_{b \in B} d(a, b), 
    \sup_{b \in B} \inf_{a \in A} d(a, b) 
\right).
    \label{Hausdorff}
\end{equation}

This metric computes the maximum distance from any point in one set to the nearest point in the other set \cite{b21}. The Hausdorff distance is particularly suitable for this case when compared to more traditional metrics such as Mean Absolute Error (MAE) or Mean Squared Error (MSE), especially because the estimated trajectory $\boldsymbol{p'}$ and the ground truth $\boldsymbol{g}$ do not necessarily contain the same number of points. This mismatch arises because the number of experience nodes is automatically determined by the algorithm.

\begin{figure}[]
\centerline{\includegraphics[width=0.6\linewidth]{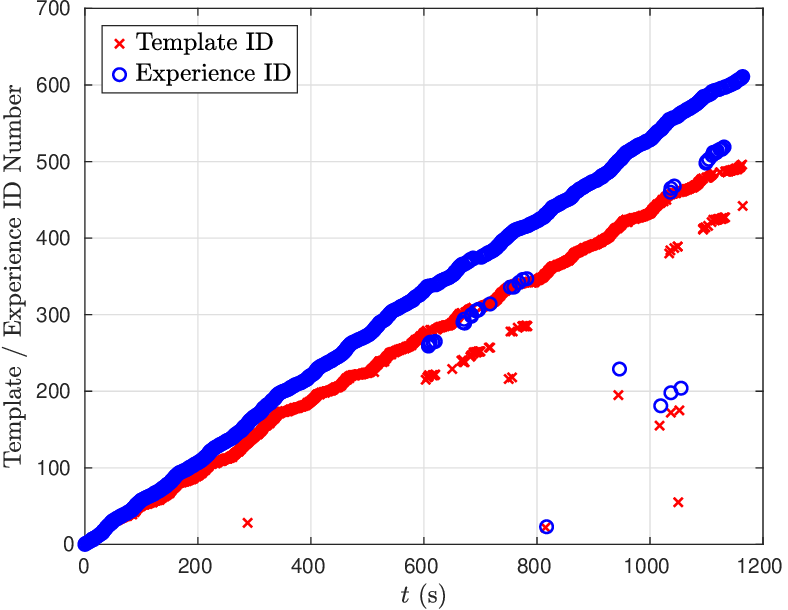}}
\caption{Active experience and visual template - MMC Lake}
\label{vt_ex_ids}
\end{figure}

For the evaluated 900-meter trajectory, the Hausdorff distance was $d_H(g, p') \approx 8.35$. This is a reasonable value if compared to the accuracy of the onboard GPS receiver (specifically, the Garmin 19X HVS) which offers an accuracy of 5–10 meters using GPS alone, and up to 3 meters with Wide Area Augmentation System (WAAS). Figure ~\ref{vt_ex_ids} illustrates the activation timeline of experiences and visual templates throughout the experiment. The upper blue line (bounding line) indicates the creation of new experience nodes, while the red line shows the creation of visual templates. Short segments below these lines indicate successful re-recognition of previously visited locations. Loop closures are marked by the start and end points of these segments.

\section{Conclusion}

This work introduced OpenRatSLAM2, a new ROS2-based version of OpenRatSLAM, designed as a modular framework for online and offline operation. Unlike navigation systems that rely on probabilistic methods, laser range sensors, or occupancy maps, RatSLAM works with only a low-resolution monocular camera. It also supports integration of odometric data from additional sensors, such as encoders and IMUs. To the best of our knowledge, this is the first work detailing the application of RatSLAM to an unmanned surface vehicle. Field experiments demonstrated that this approach can localize an USV in GPS-denied scenarios with an error margin acceptable for robotic applications. Furthermore, we analyzed the influence of algorithm parameterization, identified the most critical ones for performance, and provided recommended default values.
Future work will explore autonomous or supervised optimization of OpenRatSLAM2’s local view and pose cell parameters, using predefined dataset segments with known loop closures. 

\section*{Acknowledgment}
This research was supported in part by NSF grants IIS-2024733 and IIS-2331908, the Office of Naval Research grant N00014-23-1-2789, the U.S. Department of Defense grant 78170-RT-REP, and the Florida Department of Environmental Protection grant INV31.


\begin{thebibliography}{00}
\bibitem{b1} G. Xia and G. Wang,  ``INS/GNSS Tightly-Coupled Integration Using Quaternion-Based AUPF for USV''. Sensors, 2016, 16, 1215.

\bibitem{newaz2023}
A.~A.~R. Newaz, P.~Padrao, J.~Fuentes, T.~Alam, G.~Govindarajan, and L.~Bobadilla, 
``LCD-RIG: Limited Communication Decentralized Robotic Information Gathering Systems,'' 
\textit{IEEE Robotics and Automation Letters}, 
vol.~9, no.~11, pp.~10034--10041, 2024, 

\bibitem{puthumanaillam2025}
G. Puthumanaillam, P. Padrao, J. Fuentes, P. Thangeda, W. E. Schafer, J. H. Song, K. Jagdale, L. Bobadilla, and M. Ornik, 
``TRACE: A Self-Improving Framework for Robot Behavior Forecasting with Vision-Language Models,'' 
\textit{arXiv preprint arXiv:2503.00761}, 2025. [Online]. 


\bibitem{b2} M.C. Menezes , M.E.S. Munoz, E.P. Freitas , S. Cheng, T. Walther, A.A. Neto, P.R.A. Ribeiro and A.C.M. Oliveira, ``Automatic Tuning of RatSLAM's Parameters by Irace and Iterative Closest Point,'' (2020) IECON Proceedings (Industrial Electronics Conference), 2020, art. no. 9254718, pp. 562 - 568.

\bibitem{b3} W. Liu, Y. Liu and R. Bucknall, ``A Robust Localization Method for Unmanned Surface Vehicle (USV) Navigation Using Fuzzy Adaptive Kalman Filtering,'' in IEEE Access, vol. 7, pp. 46071-46083, 2019.


\bibitem{b4} C. Hide, M. Terry and S. Martin, ``Adaptive Kalman filtering for low-cost INS/GPS''. The Journal of Navigation, 2003. 56. 143 - 152. 10.1017/S0373463302002151.

\bibitem{b5} X. Liu, Z. Hu, Z. Sun, J. Lu, W. Xie and W. Zhang, "A VIO-Based Localization Approach in GPS-denied Environments for an Unmanned Surface Vehicle," 2023 International Conference on Advanced Robotics and Mechatronics (ICARM), Sanya, China, 2023, pp. 912-917.


\bibitem{b6} J. Burbank, T. Greene and N. Kaabouch, “Detecting and Mitigating Attacks on GPS Devices,” Sensors, vol. 24, no. 17, art. no. 5529, 2024.

\bibitem{b7} J. Xu, N. Yan and F. Tang, ``An Improvement of Loop Closure Detection Based on BoW for RatSLAM,'' 2022 37th Youth Academic Annual Conference of Chinese Association of Automation (YAC), Beijing, China, 2022, pp. 634-639.


\bibitem{b8} C.A.P. Pizzino,  R.R. Costa, D. Mitchell and  P.A. Vargas, ``NeoSLAM: Long-Term SLAM Using Computational Models of the Brain''. Sensors, 2024, 24, 1143. 

\bibitem{b9} M. J. Milford and G. F. Wyeth, ``Mapping a Suburb With a Single Camera Using a Biologically Inspired SLAM System,'' in IEEE Transactions on Robotics, vol. 24, no. 5, pp. 1038-1053, Oct. 2008.

\bibitem{b10} M.E. Muñoz, M. C. Menezes, E. Freitas, S. Cheng, P.R. Ribeiro,  A. Neto and  A.C. Oliveira, ``xRatSLAM: An Extensible RatSLAM Computational Framework,'' Sensors 2022, 22, 8305.

\bibitem{b11} D. Ball, S. Heath, M. Milford, G. Wyeth and J Wiles,
``A navigating rat animat'', Artificial Life XII: Proceedings of the 12th International Conference on the Synthesis and Simulation of Living Systems, ALIFE 2010, pp. 804 - 811.

\bibitem{b12} D. Ball, S. Heath, J. Wiles,  G. Wyeth, P. Corke and M. Milford, ``OpenRatSLAM: an open source brain-based SLAM system,'' Auton Robot 34, 149–176, 2013.

\bibitem{b14} F. Yu, J. Shang, Y. Hu and M. Milford. ``NeuroSLAM: a brain-inspired SLAM system for 3D environments'', Biol Cybern, 2019, Volume 113, pages 515–545.

\bibitem{b15} L. Silveira, F. Guth, P. Drews-Jr, P. Ballester, M. Machado, F. Codevilla, N. Duarte-Filho, S. Botelho, ``An Open-source Bio-inspired Solution to Underwater SLAM'', IFAC 2015, Pages 212-217.

\bibitem{b16} B. Manuel Pirozzo, M. De Paula, S. Aldo Villar and G. Gabriel Acosta, ``Underwater Rat-SLAM with Memristive Spiking Neural Networks,'' OCEANS 2024 - Halifax, Halifax, NS, Canada, 2024, pp. 1-8.

\bibitem{b17} M. J. Milford, G. F. Wyeth and D. Prasser, ``RatSLAM: a hippocampal model for simultaneous localization and mapping,'' IEEE International Conference on Robotics and Automation, 2004. Proceedings. ICRA '04. 2004, New Orleans, LA, USA, 2004, pp. 403-408 Vol.1.

\bibitem{b18} M. C. Menezes et al., ``Automatic Tuning of RatSLAM’s Parameters by Irace and Iterative Closest Point,'' IECON 2020 The 46th Annual Conference of the IEEE Industrial Electronics Society, Singapore, 2020, pp. 562-568.

\bibitem{b18_2} J. S. Taube, R. U. Muller and J. B. Ranck Jr, ``Head-direction cells recorded from the postsubiculum in freely moving rats. I. Description and quantitative analysis'', 1990, The Journal of neuroscience : the official journal of the Society for Neuroscience. 10. 420-35. 

\bibitem{b18_3} J. S. Taube, R. U. Muller and J. B. Ranck Jr, ``Head-direction cells recorded from the postsubiculum in freely moving rats. II. Effects of environmental manipulations'', 1990, The Journal of neuroscience : the official journal of the Society for Neuroscience. 10(2). 436-47.

\bibitem{b19} M. J. Milford and G. F. Wyeth, ``Persistent Navigation and Mapping using a Biologically Inspired SLAM System'', The International Journal of Robotics Research, 2009, 29(9), pp. 1131-1153.

\bibitem{b20} M. Milford, G. Wyeth and D. Prasser, ``RatSLAM on the Edge: Revealing a Coherent Representation from an Overloaded Rat Brain'', 2006 ,IEEE/RSJ International Conference on Intelligent Robots and Systems, Beijing, China, 2006, pp. 4060-4065.


\bibitem{b21} B. Sendov, ``Hausdorff Distance. In: Beer, G. (eds) Hausdorff Approximations.'', 1990, Mathematics and Its Applications', vol 50, Springer, Dordrecht.


\end{thebibliography}
\end{document}